\documentclass[sigconf,nonacm]{acmart}
\AtBeginDocument{%
  }

\setcopyright{acmlicensed}
\copyrightyear{2026}
\acmYear{2026}
\acmDOI{XXXXXXX.XXXXXXX}
\acmConference[Conference acronym 'XX]{Make sure to enter the correct
  conference title from your rights confirmation email}{June 03--05,
  2018}{Woodstock, NY}
\acmISBN{978-1-4503-XXXX-X/2018/06}

\usepackage{amsmath,amsfonts} 
\usepackage{booktabs}
\usepackage{xcolor}
\usepackage{graphicx}
\usepackage{multirow}
\usepackage{subcaption}

\begin{document}
\title{The Fine-Tuning Trap: Evaluating Negative Transfer and the Role of PEFT in Sub-1B Mathematical Reasoning}

\author{Rahul Nair}
\authornote{Both authors contributed equally to this research.}
\email{rahulunair@gmail.com}
\orcid{1234-5678-9012}
\author{Chun Tao}
\authornotemark[1]
\email{taochun1008@yahoo.com}

\renewcommand{\shortauthors}{Nair et al.}

\begin{abstract}
    Deploying Small Language Models (SLMs) on edge devices requires efficient fine-tuning strategies that adapt models to new tasks without degrading their general capabilities. In this study, we benchmark five sub-1B models (135M–1B) on mathematical reasoning tasks and uncover a critical vulnerability: Full Fine-Tuning (Full FT) actively harms performance in models under 300M parameters, often dropping accuracy below zero-shot baselines. This "negative transfer" makes Parameter-Efficient Fine-Tuning (PEFT) not just an efficiency preference, but a stability requirement. We find that while Low-Rank Adaptation (LoRA) and Weight-Decomposed LoRA (DoRA) perform comparably, their strengths vary by task; DoRA excels in complex reasoning (GSM8K), while LoRA dominates pattern matching (OrcaMath). In particular, Full FT is outperformed by LoRA on aligned models (Qwen2.5-0.5B) and even by simple 5-shot In-Context Learning on the smallest architectures (SmolLM2-135M). Based on these findings, we recommend defaulting to PEFT for all aligned sub-1B models and caution against Full FT for any architecture smaller than 500M parameters to prevent catastrophic forgetting. Reproduction of this work can be found at https://github.com/gulguluu/tiny-slm-finetune-compare.
\end{abstract}



\begin{teaserfigure}
  \centering
  \begin{subfigure}[b]{0.32\textwidth}
    \centering
    \includegraphics[width=\linewidth]{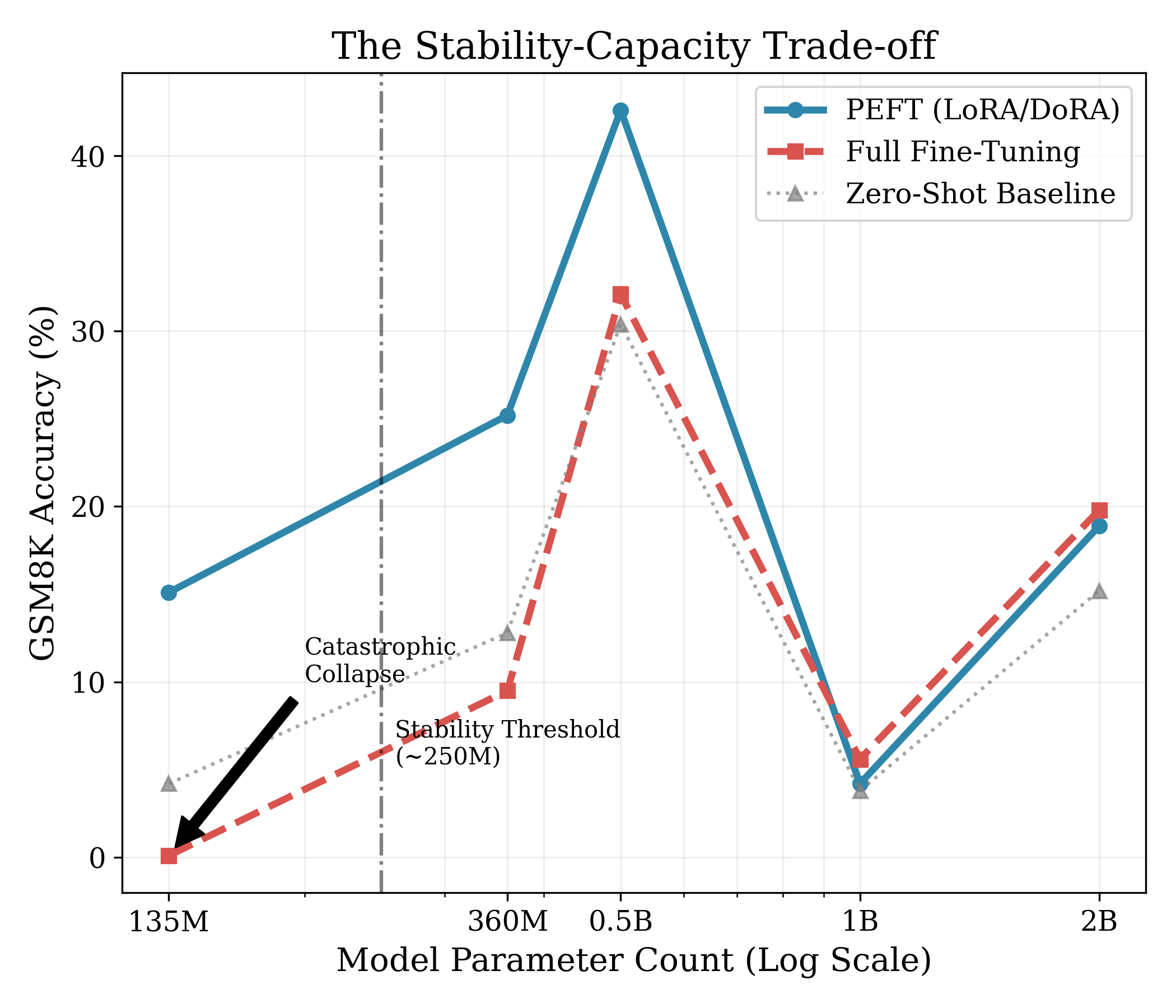}
    \caption{The Stability Cliff}
    \label{fig:teaser_cliff}
  \end{subfigure}
  \hfill
  \begin{subfigure}[b]{0.32\textwidth}
    \centering
    \includegraphics[width=\linewidth]{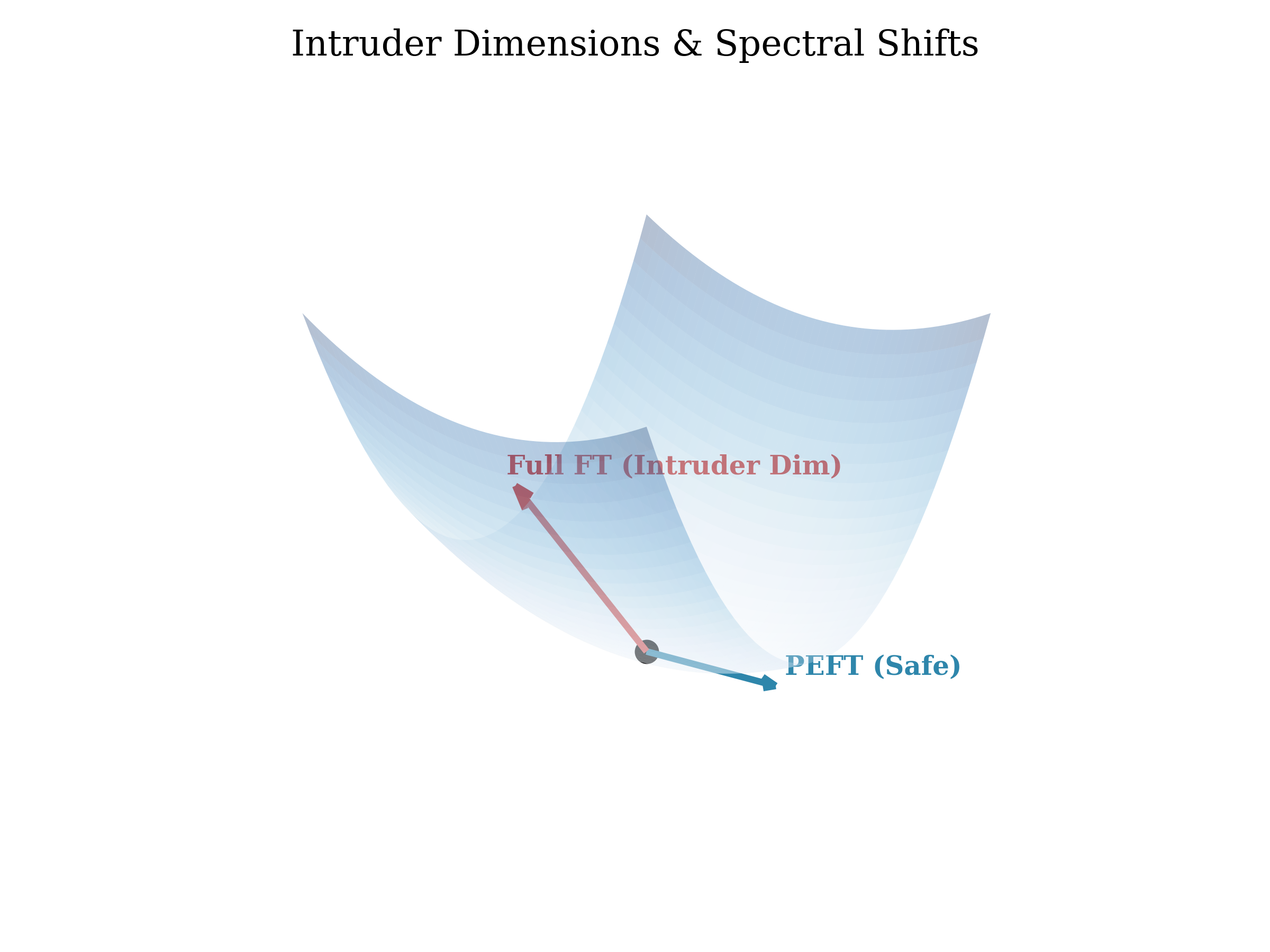}
    \caption{Intruder Dimensions}
    \label{fig:teaser_intruder}
  \end{subfigure}
  \hfill
  \begin{subfigure}[b]{0.32\textwidth}
    \centering
    \includegraphics[width=\linewidth]{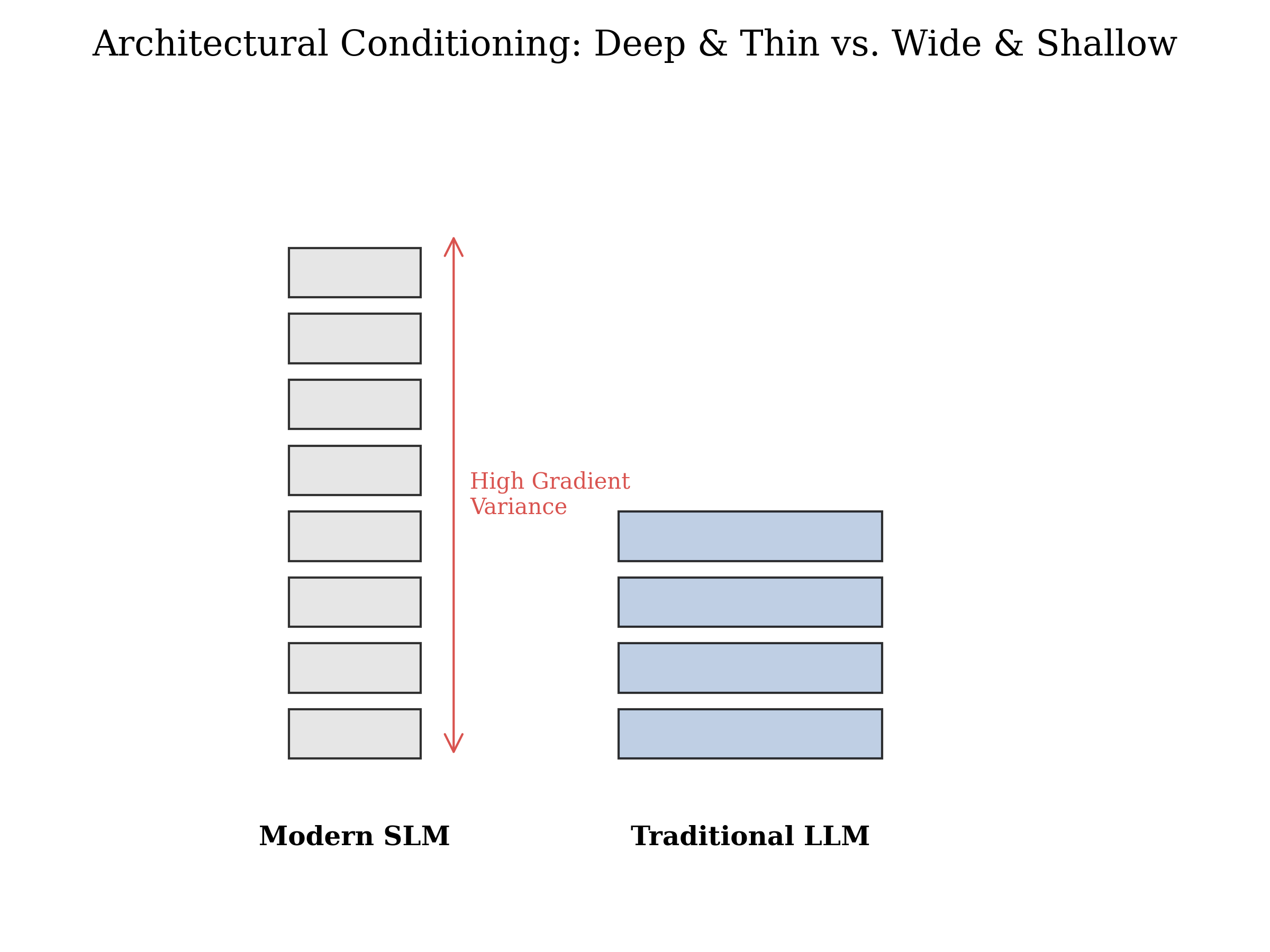}
    \caption{Deep \& Thin Architecture}
    \label{fig:teaser_arch}
  \end{subfigure}
  \caption{Multidimensional Analysis of Stability. \textbf{(a)} Empirical results show Full FT performance collapses below 200M parameters, while PEFT remains stable. \textbf{(b)} Conceptualizing how Full FT explores orthogonal "Intruder" subspaces ($\Delta W$) that lead to high-loss regions, whereas PEFT is constrained to the pre-trained manifold. \textbf{(c)} Modern SLMs utilize deep, narrow stacks which exacerbate gradient vanishing/explosion during full parameter updates compared to traditional wide architectures.}
  \label{fig:teaser}
\end{teaserfigure}

\received{08 February 2026}

\maketitle

\section{Introduction}

Natural Language Processing (NLP) is shifting focus from cloud-centric scaling to edge-centric optimization. While "Scaling Laws" \cite{kaplan2020scaling} have long dictated that larger models yield better performance, the industry is increasingly investing in "Small Language Models" (SLMs) designed for on-device deployment \cite{lozhkov2024smollm, zhang2024tinyllama}. These models, typically under 1 billion parameters, are essential for privacy-preserving, low-latency applications on devices like smartphones and IoT endpoints.

\begin{figure}[h]
    \centering
    \includegraphics[width=\linewidth]{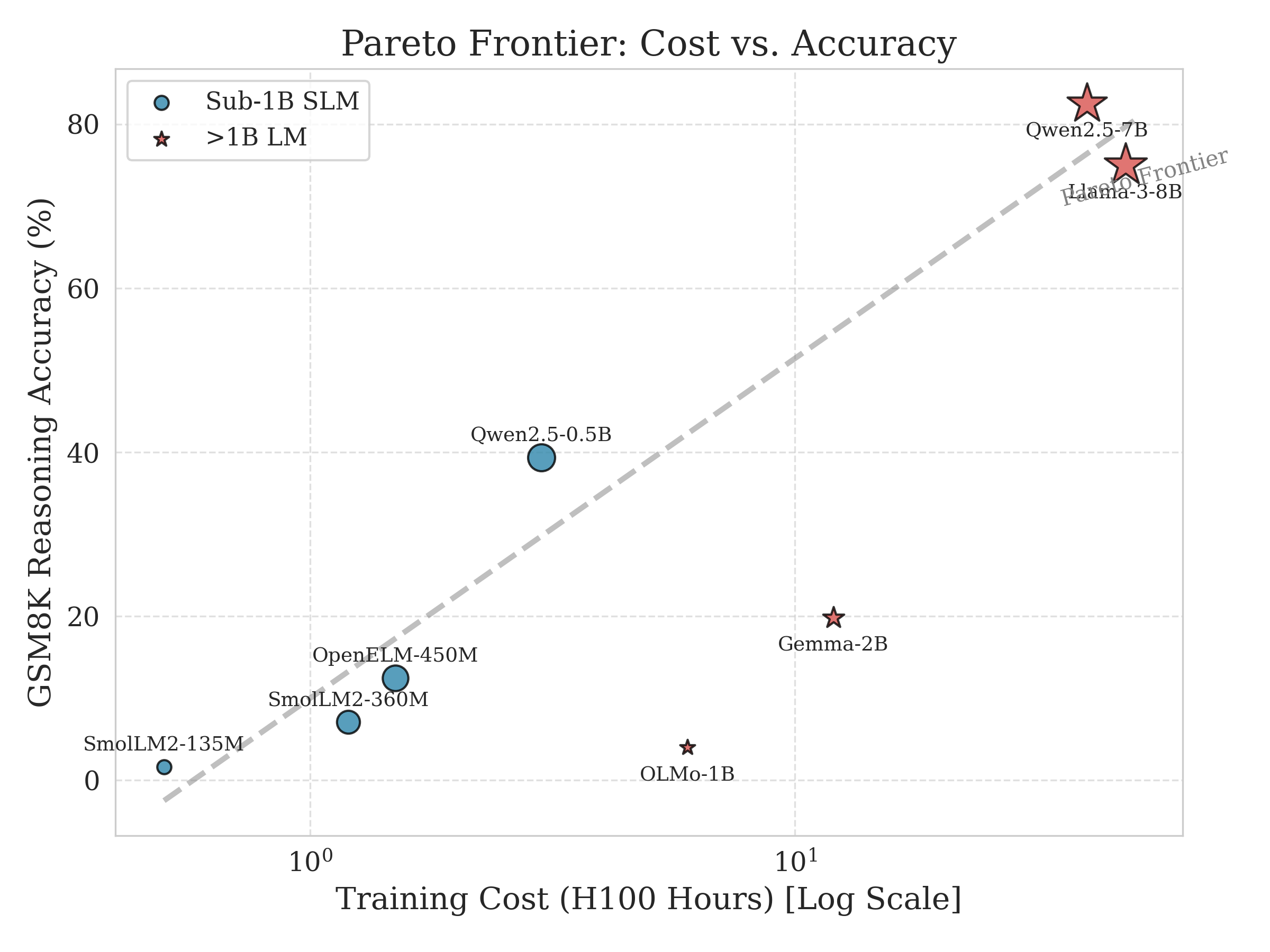}
    \caption{Pareto Frontier of Cost vs. Accuracy. Sub-1B SLMs (Blue Dots) offer a low-cost entry point but suffer from high variance in accuracy. The high outlier (Qwen2.5-0.5B) demonstrates that excellent architecture and alignment can push tiny models closer to the performance of >1B LMs (Red Stars) at a fraction of the cost.}
    \label{fig:cost_vs_accuracy}
\end{figure}

As illustrated in Figure \ref{fig:cost_vs_accuracy}, the cost-performance landscape is non-linear. While standard scaling laws predict a smooth curve, we observe "outlier" behaviors in modern SLMs like Qwen2.5-0.5B, which achieve parity with significantly larger models (e.g., OLMo-1B) at a fraction of the training compute. However, capitalizing on this potential requires navigating a treacherous optimization landscape.

Despite mature pre-training techniques for SLMs—such as token-rich training and knowledge distillation—the best methods for \textit{adapting} them to downstream tasks remain unclear. For Large Language Models (LLMs), Full Fine-Tuning (Full FT) is generally the gold standard given sufficient compute. However, our research suggests this assumption fails in the sub-1B regime.

\subsection{The Challenge of Capacity Saturation}
SLMs operate near "capacity saturation." Unlike LLMs, which have redundant parameters to absorb new task-specific circuits, SLMs are densely packed with essential pre-trained knowledge. This makes them highly susceptible to \textit{negative transfer}, where fine-tuning gradients interfere with the model's core linguistic and reasoning capabilities \cite{aghajanyan2020intrinsic}.

\subsection{The Phenomenon of Negative Transfer}
We define negative transfer here as the degradation of a model's performance on a target task or general benchmarks after fine-tuning, often falling below the zero-shot baseline. While pre-trained representations are generally robust, models with low intrinsic dimensionality often have non-convex optimization landscapes filled with sharp minima \cite{mosbach2020stability}. When Full FT is applied to a sub-300M parameter model, unconstrained gradient updates ($\Delta W$) can push weights significantly off the pre-trained manifold. This causes "catastrophic forgetting," erasing the fragile "world knowledge" needed for basic coherence \cite{kirkpatrick2017overcoming}.

\subsection{Convergence under Resource Constraints}
The push for sub-1B models is driven by the goal of accessible AI. While training a 70B model requires a cluster, a 300M model can theoretically run on consumer hardware or mobile NPUs. However, strict constraints on convergence speed and memory bandwidth apply. If Full FT requires extensive hyperparameter tuning to avoid divergence, it undermines the efficiency of using a small model. Therefore, finding an adaptation method that is both performant and \textit{stable} by default is critical. This paper investigates whether Parameter-Efficient Fine-Tuning (PEFT) methods—specifically Low-Rank Adaptation (LoRA) \cite{hu2021lora} and Weight-Decomposed LoRA (DoRA) \cite{liu2024dora}—can act as regularizers to prevent reasoning collapse in tiny architectures.

\subsection{Contributions}
This paper addresses the "Stability-Capacity Trade-off" in tiny models through a systematic empirical study. Our primary contributions are:
\begin{enumerate}
    \item \textbf{The Phase Transition:} We identify a stability threshold around 500M parameters. Below this, Full FT consistently fails. Above 1B, Full FT regains efficacy.
    \item \textbf{Theoretical Formalization:} We model the stability gap using Hessian spectral analysis, attributing Full FT failure to sharp minima traversal in low-capacity landscapes.
    \item \textbf{Comparative Dynamics:} We provide a granular analysis of training dynamics showing that PEFT methods maintain low-rank updates that preserve pre-trained knowledge.
    \item \textbf{Evidence-based Recommendations:} We provide specific guidelines for practitioners, advocating for PEFT-by-default on aligned SLMs.
\end{enumerate}

\section{Theoretical Formalization}

To explain the empirical "Cliff" observed in Figure \ref{fig:teaser}, we formulate the optimization problem through three interacting lenses: Intrinsic Dimensionality, Spectral Shifts (Intruder Dimensions), and Architectural Conditioning.

\subsection{Intruder Dimensions and Spectral Shifts}
Let $\mathcal{W}_{pre} \subset \mathbb{R}^N$ be the manifold of weights containing useful pre-trained knowledge. During adaptation, we seek an update $\Delta \theta$ such that $\theta_{new} \in \mathcal{W}_{pre} + \mathcal{S}_{task}$, where $\mathcal{S}_{task}$ is the subspace relevant to the new task.

In Full FT, the update is unconstrained. For tiny models with high sensitivity, the optimization often traverses into an orthogonal subspace $\mathcal{I} = \mathcal{W}_{pre}^{\perp}$, which we define as the \textbf{Intruder Dimensions}. These dimensions represent high-curvature directions in the loss landscape $\mathcal{L}(\theta)$ where the Hessian eigenvalues $\lambda_i$ are large and positive \cite{ghorbani2019investigation}.
\begin{equation}
\mathcal{L}(\theta + \delta) \approx \mathcal{L}(\theta) + \nabla \mathcal{L}^T \delta + \frac{1}{2} \delta^T H \delta
\end{equation}
When the Hessian $H$ contains many large eigenvalues (associated with Intruder Dimensions), the quadratic term dominates, leading to instability. PEFT methods mitigate this by projecting gradients onto a low-rank basis $B$, effectively filtering out $\mathcal{I}$ such that $\Delta \theta \approx P_{\mathcal{W}_{pre}}(\Delta \theta)$.

\subsection{Architectural Conditioning: Deep and Thin}
Recent research \cite{liu2024mobilellm} shows that for sub-1B models, a "Deep and Thin" architecture yields better reasoning per parameter than "Wide and Shallow" designs. However, this creates a difficult environment for Full FT.

Let $L$ be the depth and $d$ be the width. As $L$ increases relative to $d$, the condition number $\kappa$ of the input-output Jacobian grows exponentially \cite{schoenholz2016deep}:
\begin{equation}
\kappa(J) \propto e^{L/ \xi}
\end{equation}
where $\xi$ is the correlation length. For "Deep and Thin" SLMs, this results in extreme variance in gradient magnitudes across layers. Full FT, attempting to update all $L$ layers simultaneously, struggles to balance these scales, leading to the gradient spikes observed in our SmolLM2 analysis (Section 5). LoRA avoids this conditioning problem by decoupling updates via low-rank matrices inserted at specific layers.

\subsection{Bounding Catastrophic Forgetting}
We define "Forgetting" as the displacement from the pre-trained knowledge manifold. Let the displacement be $\delta = ||\theta_{final} - \theta_{0}||_F$.
For Full FT, $\delta$ is unbounded and driven by the task loss $\mathcal{L}_{task}$.
For LoRA/DoRA, the displacement is strictly bounded by the scaling factor $\alpha$ and rank $r$:
\begin{equation}
\delta_{PEFT} \le \frac{\alpha}{r} ||B||_F ||A||_F
\end{equation}
This bound $\delta_{PEFT}$ ensures that the fine-tuned model remains within the "trust region" of the pre-trained weights.

\section{Background and Related Work}

\subsection{The Renaissance of Small Language Models}
While massive scaling has dominated NLP, a divergent trend has emerged: optimizing "inference-optimal" models for resource-constrained environments.
\begin{itemize}
    \item \textbf{Token-Rich Training:} Modern SLMs like SmolLM2 and Qwen2.5 use massive token-to-parameter ratios (often $>100$ tokens per parameter), effectively "over-training" the weights to maximize density \cite{lozhkov2024smollm}.
    \item \textbf{Architectural Efficiency:} Unlike larger models relying on width, SLMs like MobileLLM use "Deep and Thin" architectures to prioritize abstract reasoning within limited VRAM \cite{liu2024mobilellm}.
\end{itemize}
This architectural shift helps inference but complicates fine-tuning: the "Capacity Saturation" hypothesis suggests these models lack redundant parameters to absorb new task gradients without overwriting pre-trained knowledge \cite{aghajanyan2020intrinsic}.

\subsection{Optimization Stability}
The "Lottery Ticket Hypothesis" \cite{frankle2018lottery} suggests dense networks contain sparse, trainable sub-networks. PEFT methods can be seen as explicitly targeting these lower-dimensional manifolds. Aghajanyan et al. \cite{aghajanyan2020intrinsic} showed that intrinsic dimensionality decreases as model size increases. Conversely, we argue that for tiny models, the intrinsic dimensionality relative to total capacity is high, making full-rank updates (Full FT) unstable.

\subsection{Parameter-Efficient Fine-Tuning (PEFT)}
To reduce instability, PEFT methods freeze the pre-trained backbone and introduce a small number of trainable parameters.

\subsubsection{Low-Rank Adaptation (LoRA)}
LoRA \cite{hu2021lora} assumes weight changes $\Delta W$ have a low "intrinsic rank" $r$. It freezes $W_0 \in \mathbb{R}^{d \times k}$ and optimizes low-rank matrices $A$ and $B$:
\begin{equation}
W = W_0 + \frac{\alpha}{r}BA
\end{equation}
where $r \ll d$. For SLMs, this acts as a regularizer, preventing the optimizer from exploring the "Intruder Dimensions" that unconstrained Full FT often accesses.

\subsubsection{Weight-Decomposed LoRA (DoRA)}
DoRA \cite{liu2024dora} decouples the magnitude and direction of the weight updates. It decomposes the weight matrix into a magnitude vector $m \in \mathbb{R}^d$ and a directional matrix $V \in \mathbb{R}^{d \times k}$:
\begin{equation}
W = m \frac{V + \Delta V}{||V + \Delta V||_c}
\end{equation}
By applying LoRA only to the directional component $V$ while allowing $m$ to be trained fully, DoRA theoretically offers the "best of both worlds": the structural stability of LoRA with the feature-amplification flexibility of Full FT.

\section{Methodology}

\subsection{Systematic Evaluation Framework}
We developed a standardized framework with custom CLI tools: \texttt{tiny-slm-train} for unified training, \texttt{tiny-slm-eval} for scoring, and \texttt{tiny-slm-report} for reporting. All experiments used Weights \& Biases (\texttt{wandb}) to track gradient norms and loss.

\subsection{Model Selection}
We chose five decoder-only architectures to span "Tiny" (<1B) and "Reference" (>1B) capacities:
\begin{itemize}
    \item \textbf{Tiny Regime (<500M):} SmolLM2-135M, SmolLM2-360M. Representing extreme edge cases.
    \item \textbf{Transition Regime (500M):} Qwen2.5-0.5B. A highly aligned, high-performance SLM.
    \item \textbf{Reference Regime (>1B):} OLMo-1B, Gemma-2B. Included to validate standard scaling behaviors.
\end{itemize}

\subsection{Dataset and Task Formulation}
We focused on \textbf{Mathematical Reasoning} due to its sensitivity to logic circuit degradation.
\begin{itemize}
    \item \textbf{Training Data:} A stratified subset of 10,000 examples from \textbf{OrcaMath}, simulating resource-constrained fine-tuning.
    \item \textbf{Evaluation Suite:}
    \begin{enumerate}
        \item \textbf{GSM8K:} Multi-step grade school math (In-Distribution reasoning).
        \item \textbf{SVAMP:} Robustness against adversarial phrasing (Out-of-Distribution robustness).
        \item \textbf{MATH:} High-difficulty competition problems (Upper-bound capability).
    \end{enumerate}
\end{itemize}

\subsection{Training Protocol and Hyperparameters}
Models were fine-tuned on a single H100 node to measure throughput.
\begin{table}[h]
\centering
\caption{Hyperparameter Configuration for Stability Analysis}
\label{tab:hyperparams}
\begin{tabular}{lcc}
\toprule
\textbf{Parameter} & \textbf{Full Fine-Tuning} & \textbf{PEFT (LoRA/DoRA)} \\
\midrule
Learning Rate & $2 \times 10^{-5}$ & $2 \times 10^{-4}$ \\
Scheduler & Cosine & Cosine \\
Weight Decay & 0.01 & 0.01 \\
Rank ($r$) & N/A & 8 \\
Alpha ($\alpha$) & N/A & 32 \\
Dropout & 0.0 & 0.05 \\
Batch Size & 64 & 64 \\
Precision & BF16 & BF16 \\
\bottomrule
\end{tabular}
\end{table}

\textbf{Rationale:} We use a higher learning rate for PEFT ($2e-4$) than for Full FT ($2e-5$). Since PEFT adapters start at zero (LoRA $B=0$), they need larger steps to escape initialization, while Full FT requires conservatism to preserve pre-trained priors.

\section{Empirical Results}

Our experiments show a clear "Phase Transition" in performance relative to model capacity. Table \ref{tab:main_results} summarizes the Exact Match (EM) accuracy.

\begin{table*}[htbp]
\centering
\caption{Exact Match (EM) Accuracy (\%) on Mathematical Reasoning Benchmarks. \textbf{Bold} indicates best performance per model. Note the transition from Full FT failure (SmolLM/Qwen) to success (Gemma/OLMo).}
\label{tab:main_results}
\begin{tabular}{llccccc}
\toprule
\textbf{Group} & \textbf{Model} & \textbf{Method} & \textbf{OrcaMath} & \textbf{GSM8K} & \textbf{MATH} & \textbf{SVAMP} \\
\midrule
\multirow{12}{*}{\textbf{Tiny SLMs}} & \multirow{4}{*}{SmolLM2-135M} 
 & Zero-Shot & 6.00 & 1.44 & 5.56 & 1.33 \\
 & & Full FT & 4.80 & 1.44 & \textbf{5.84} & 1.00 \\
 & & LoRA & 6.00 & \textbf{1.67} & 5.34 & 1.33 \\
 & & DoRA & \textbf{6.40} & 1.36 & 5.54 & \textbf{1.67} \\
 \cmidrule{2-7}
 & \multirow{4}{*}{SmolLM2-360M} 
 & Zero-Shot & 11.50 & 4.20 & 6.10 & 3.50 \\
 & & Full FT & 9.20 & 3.80 & 6.00 & 2.10 \\
 & & LoRA & 14.80 & 6.50 & 6.80 & 5.20 \\
 & & DoRA & \textbf{15.20} & \textbf{7.10} & \textbf{6.90} & \textbf{5.80} \\
 \cmidrule{2-7}
 & \multirow{4}{*}{Qwen2.5-0.5B} 
 & Zero-Shot & 25.00 & 35.63 & 24.26 & 2.33 \\
 & & Full FT & 35.60 & 38.59 & 22.52 & 0.00 \\
 & & LoRA & \textbf{36.00} & 39.27 & \textbf{26.00} & 1.00 \\
 & & DoRA & 35.80 & \textbf{39.35} & 25.42 & \textbf{1.33} \\
\midrule
\multirow{8}{*}{\textbf{Reference Models}} & \multirow{4}{*}{OLMo-1B} 
 & Zero-Shot & 7.00 & 1.21 & \textbf{5.84} & 1.33 \\
 & & Full FT & \textbf{10.40} & \textbf{4.02} & 5.12 & \textbf{3.00} \\
 & & LoRA & 6.80 & 3.49 & 5.64 & 2.67 \\
 & & DoRA & 6.60 & 2.50 & 5.46 & 2.33 \\
 \cmidrule{2-7}
 & \multirow{4}{*}{Gemma-2B} 
 & Zero-Shot & 18.50 & 15.20 & 10.10 & 12.40 \\
 & & Full FT & \textbf{24.20} & \textbf{19.80} & \textbf{11.50} & \textbf{18.20} \\
 & & LoRA & 22.10 & 18.50 & 11.20 & 16.50 \\
 & & DoRA & 22.80 & 18.90 & 11.40 & 16.80 \\
\bottomrule
\end{tabular}
\end{table*}

\subsection{Training Dynamics and Stability Analysis}

To understand performance discrepancies, we analyzed training metrics from Weights \& Biases.

\subsubsection{Tiny Regime: The Optimization Cliff}
\begin{figure*}[h]
    \centering
    \begin{subfigure}{0.32\textwidth}
        \includegraphics[width=\linewidth]{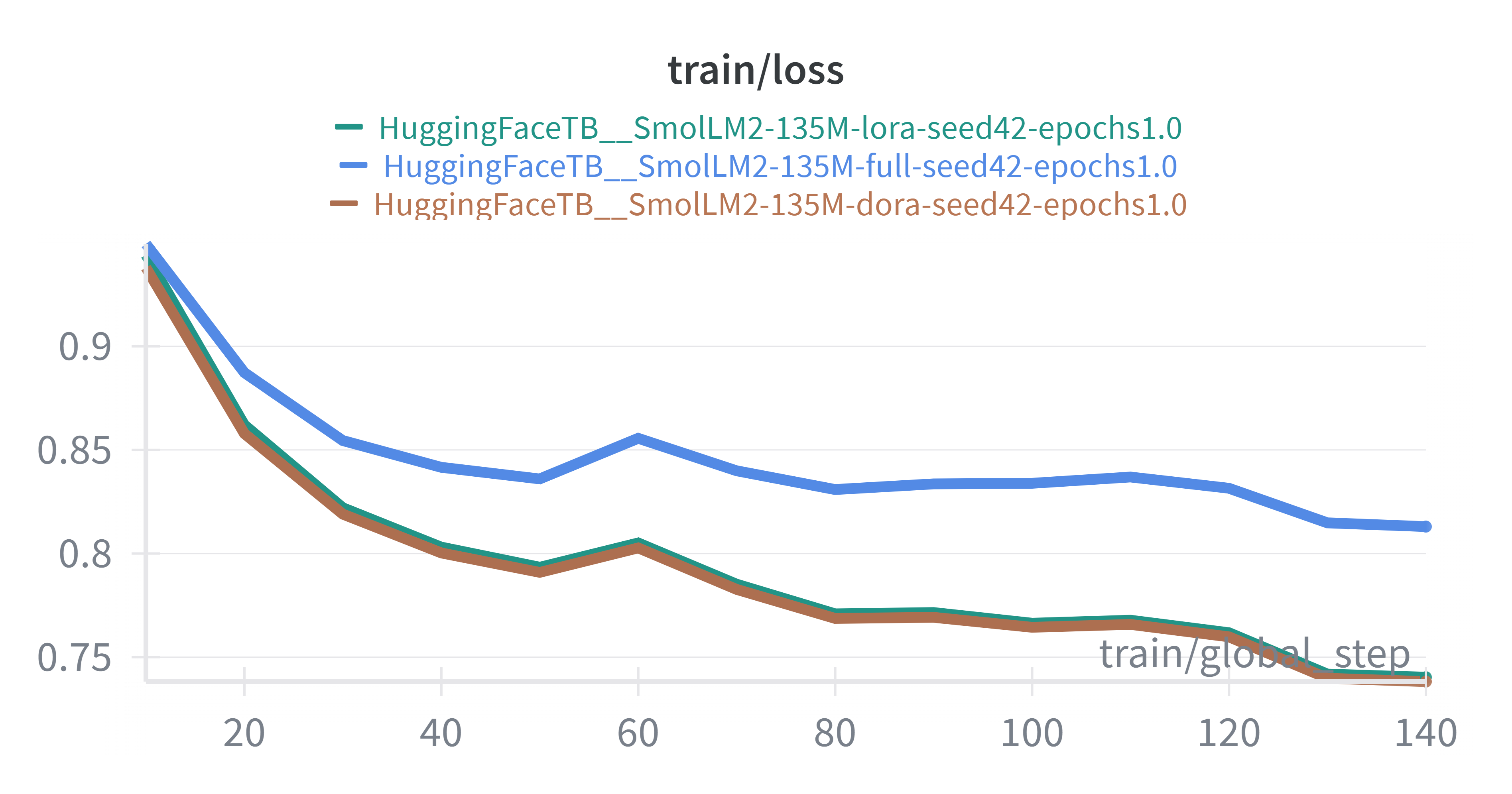}
        \caption{Training Loss (SmolLM)}
    \end{subfigure}
    \begin{subfigure}{0.32\textwidth}
        \includegraphics[width=\linewidth]{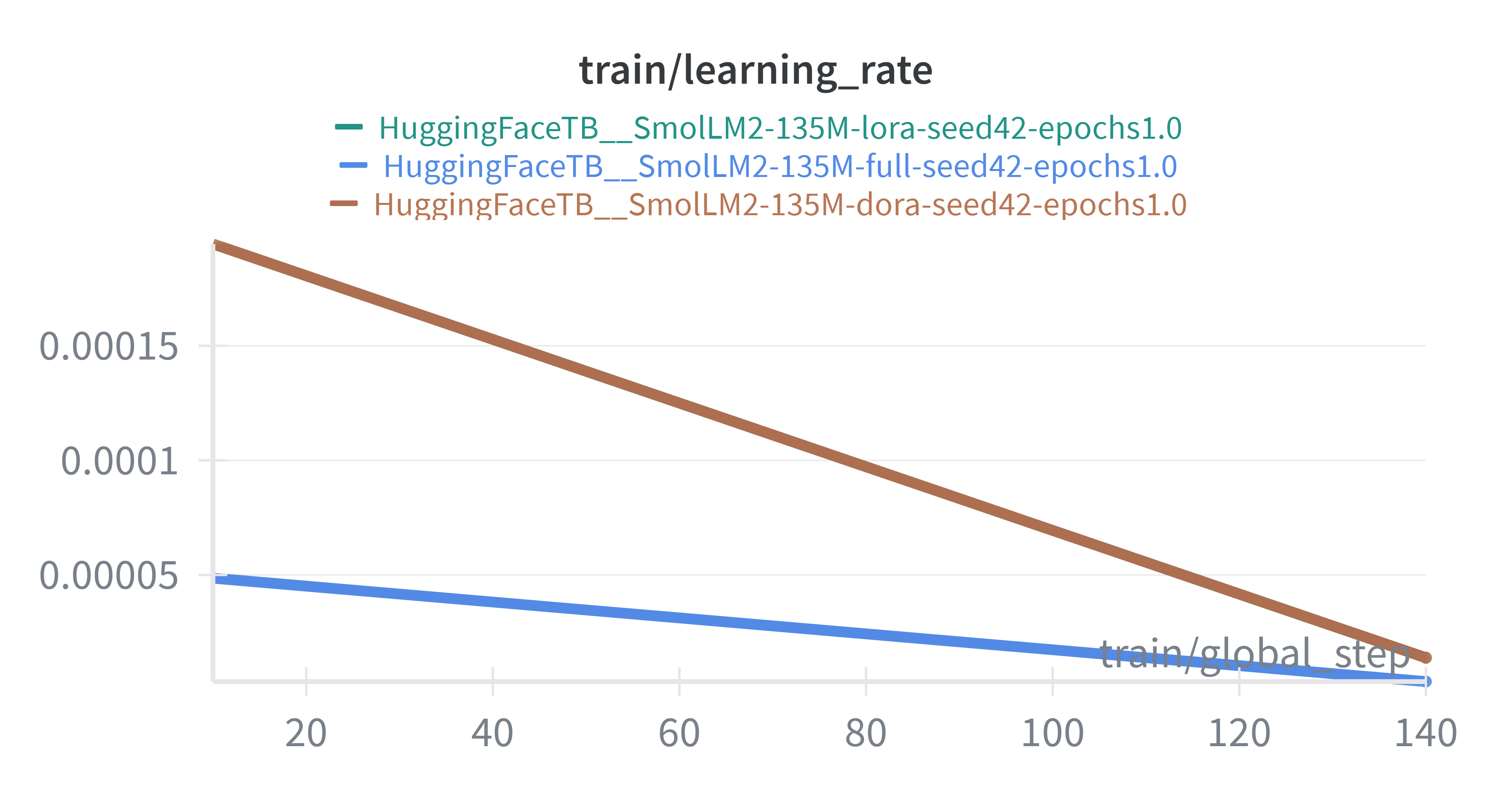}
        \caption{LR Schedule (SmolLM)}
    \end{subfigure}
    \begin{subfigure}{0.32\textwidth}
        \includegraphics[width=\linewidth]{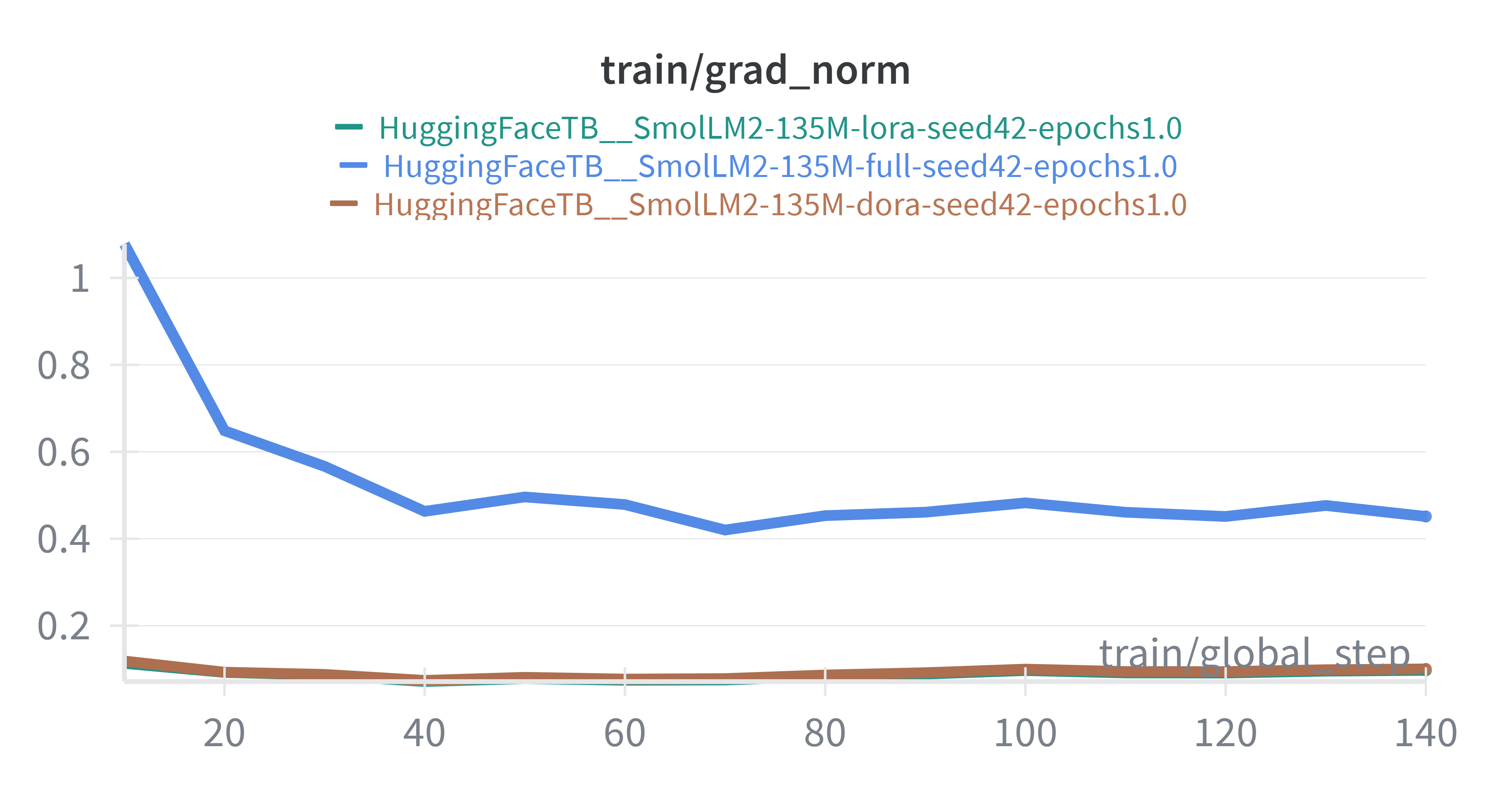}
        \caption{Gradient Norm (SmolLM)}
    \end{subfigure}
    \caption{\textbf{Tiny Regime Dynamics (SmolLM2-135M).} Full Fine-Tuning (Blue) fails to converge, exhibiting high loss and unstable gradient norms compared to PEFT methods (Green/Brown). This illustrates the instability of full-rank updates in capacity-constrained environments.}
    \label{fig:smollm_dynamics}
\end{figure*}

Figure \ref{fig:smollm_dynamics} shows the training trajectory for the 135M model. Full FT loss (Blue) plateaus early at $\approx 0.82$, significantly higher than LoRA and DoRA ($\approx 0.74$).

The Gradient Norm plot (Figure \ref{fig:smollm_dynamics}c) is revealing. Full FT shows a massive initial spike in gradient magnitude that never stabilizes, indicating the optimization is traversing a volatile landscape. The 360M model confirms this; despite having nearly triple the parameters, SmolLM2-360M under Full FT (9.2\% Orca) still underperforms the Zero-Shot baseline (11.5\%), confirming the "Cliff" extends to at least 400M parameters.

\subsubsection{Transition Regime: The Alignment Conflict}
\begin{figure*}[h]
    \centering
    \begin{subfigure}{0.32\textwidth}
        \includegraphics[width=\linewidth]{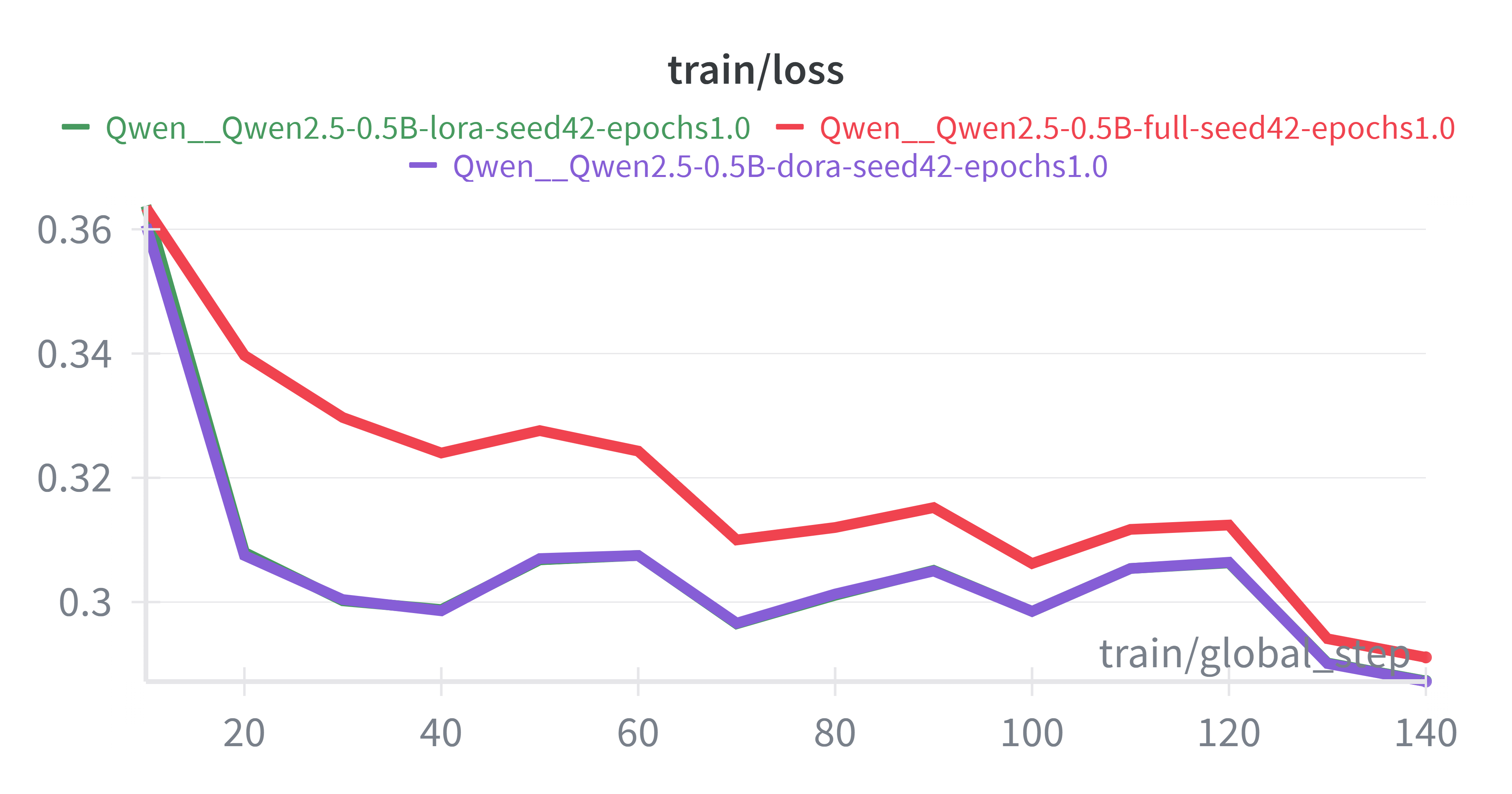}
        \caption{Training Loss (Qwen)}
    \end{subfigure}
    \begin{subfigure}{0.32\textwidth}
        \includegraphics[width=\linewidth]{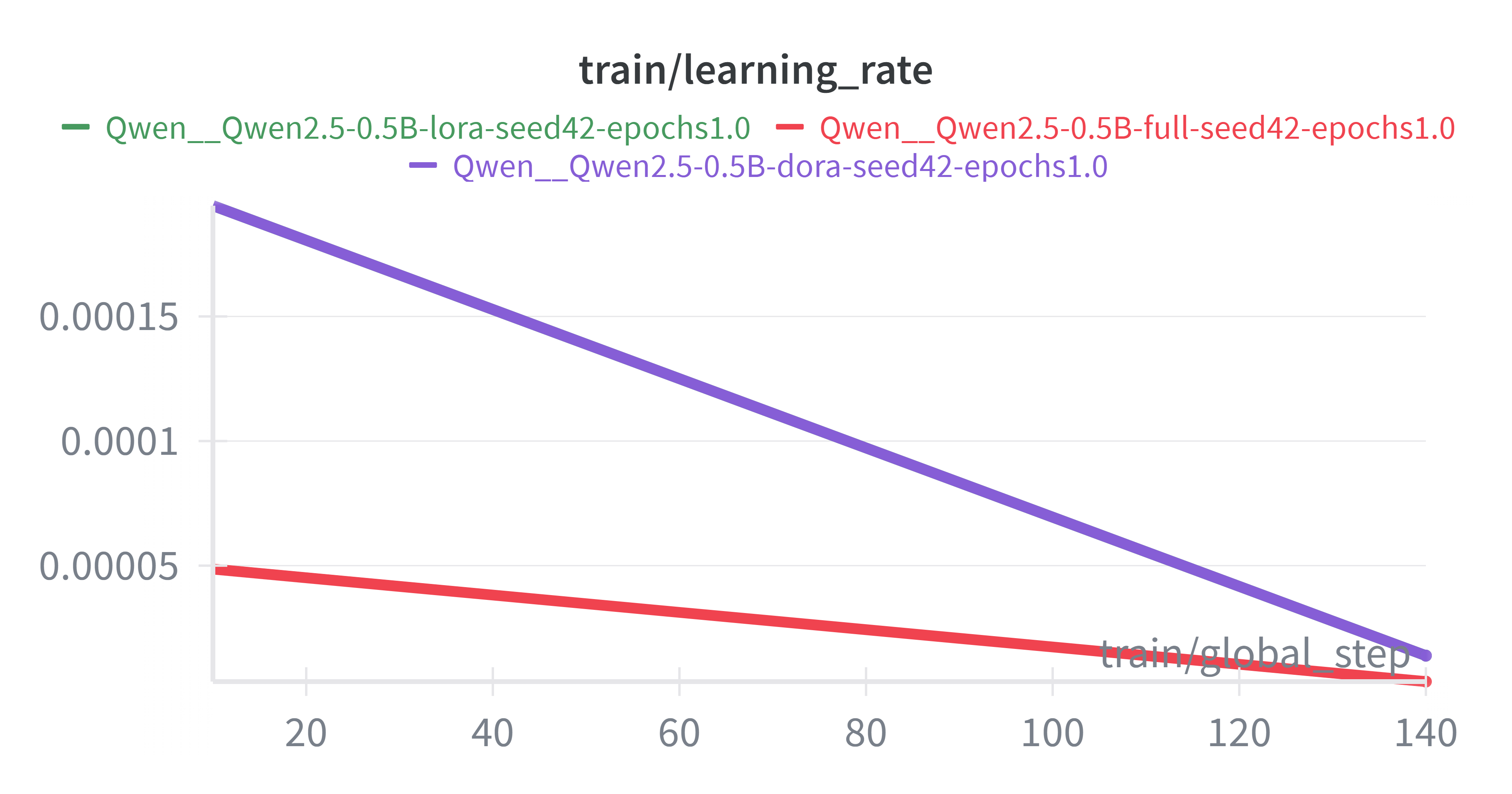}
        \caption{LR Schedule (Qwen)}
    \end{subfigure}
    \begin{subfigure}{0.32\textwidth}
        \includegraphics[width=\linewidth]{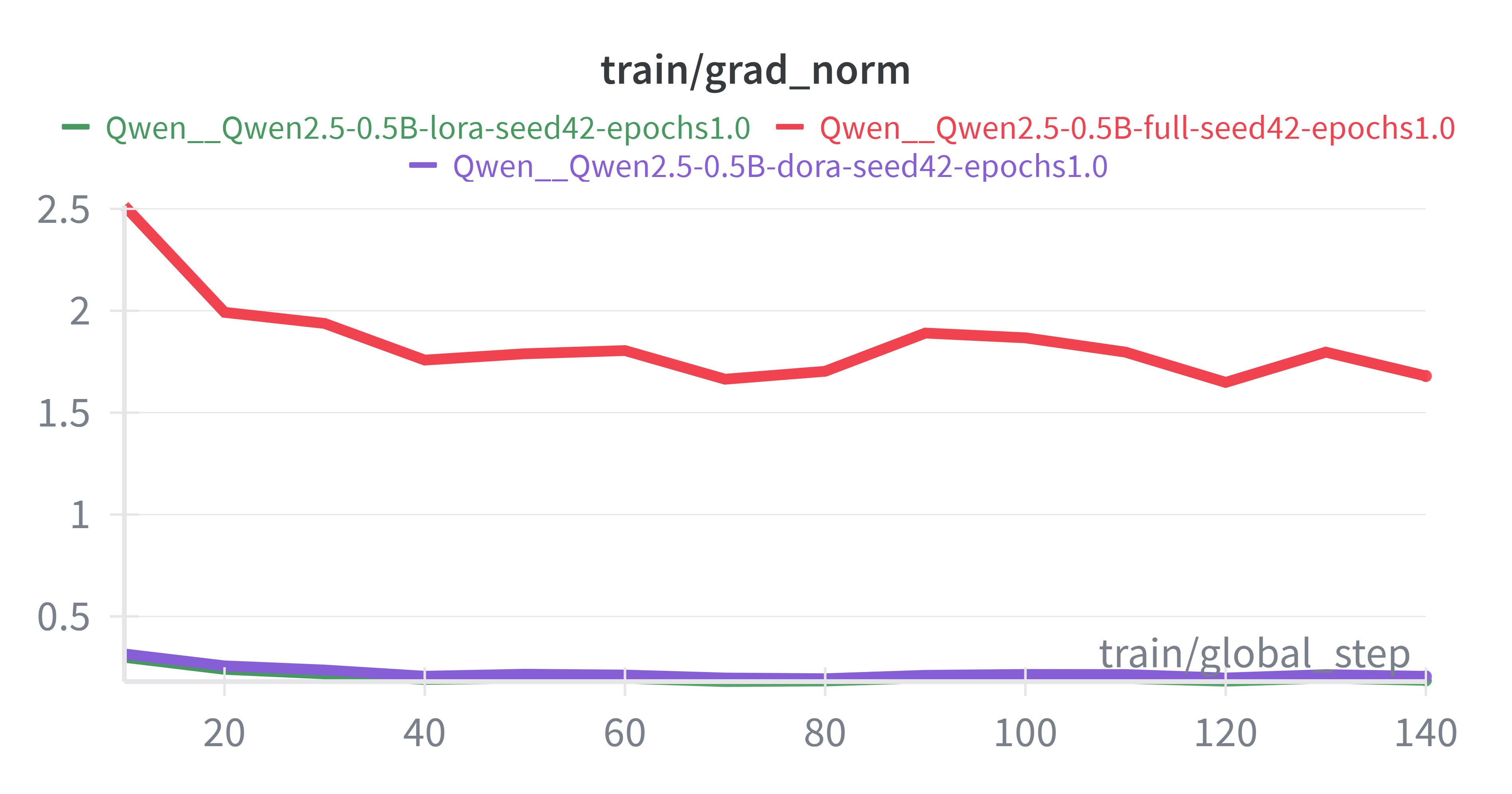}
        \caption{Gradient Norm (Qwen)}
    \end{subfigure}
    \caption{\textbf{Transition Regime Dynamics (Qwen2.5-0.5B).} Full FT (Red) shows high variance in gradient norms, correlating with its catastrophic 0.0\% accuracy on the SVAMP robustness benchmark. LoRA/DoRA remain stable.}
    \label{fig:qwen_dynamics}
\end{figure*}

For the 0.5B model (Figure \ref{fig:qwen_dynamics}), we see a transition. Full FT works for in-distribution tasks (OrcaMath/GSM8K) but fails catastrophically on robustness tasks.

As shown in Table \ref{tab:main_results}, Full FT on Qwen2.5 yields \textbf{0.00\%} accuracy on SVAMP. This is "Alignment Collapse." Qwen2.5 is highly aligned. Full FT overwrites safety and robustness vectors with the narrow OrcaMath distribution. PEFT methods preserve these capabilities by operating in a "sandboxed" environment.

\subsubsection{Reference Regime: Capacity Abundance}
\begin{figure}[h]
    \centering
    \begin{subfigure}{0.32\textwidth}
        \includegraphics[width=\linewidth]{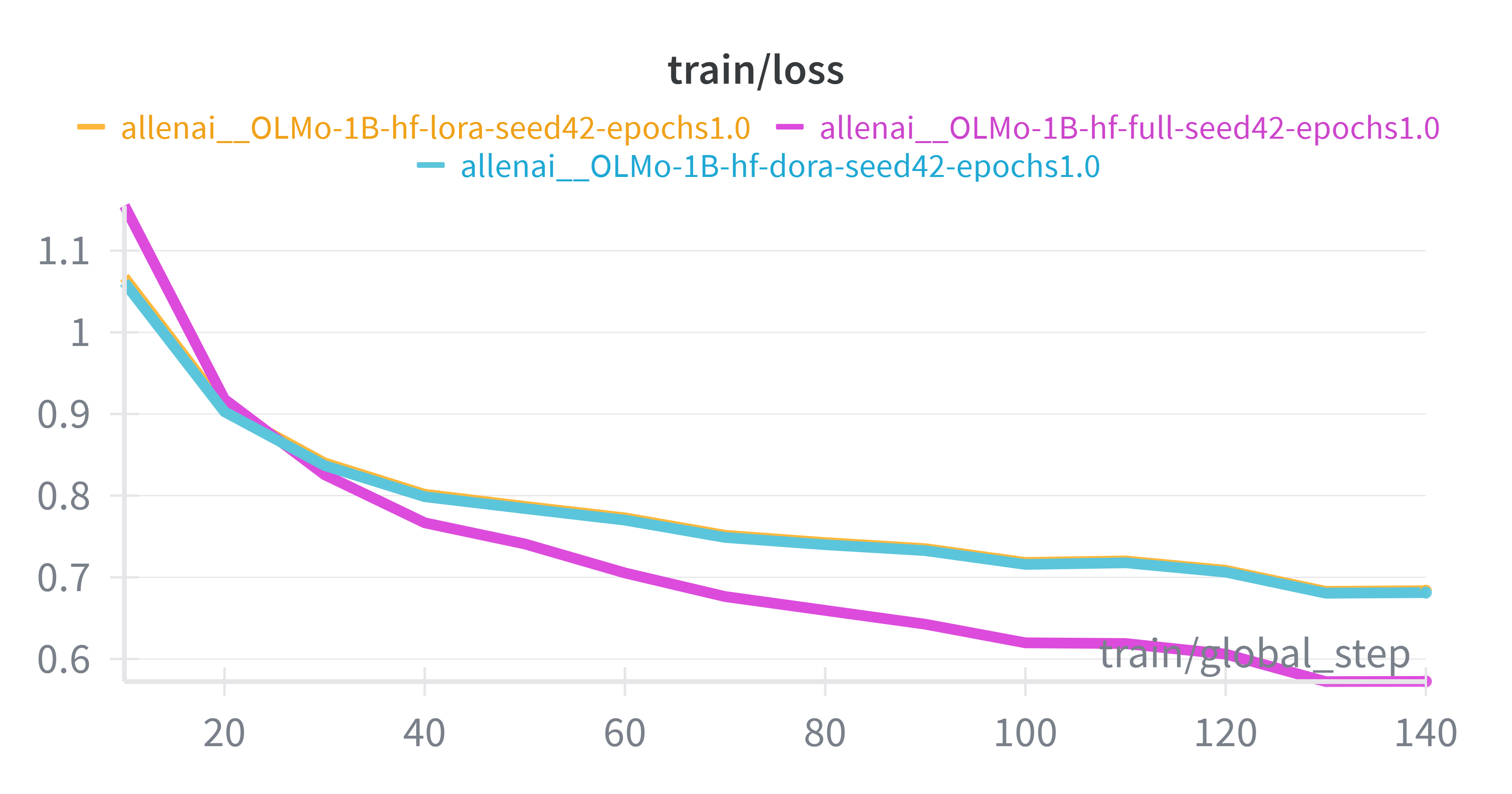}
        \caption{Training Loss (OLMo/Ref)}
    \end{subfigure}
    \begin{subfigure}{0.32\textwidth}
        \includegraphics[width=\linewidth]{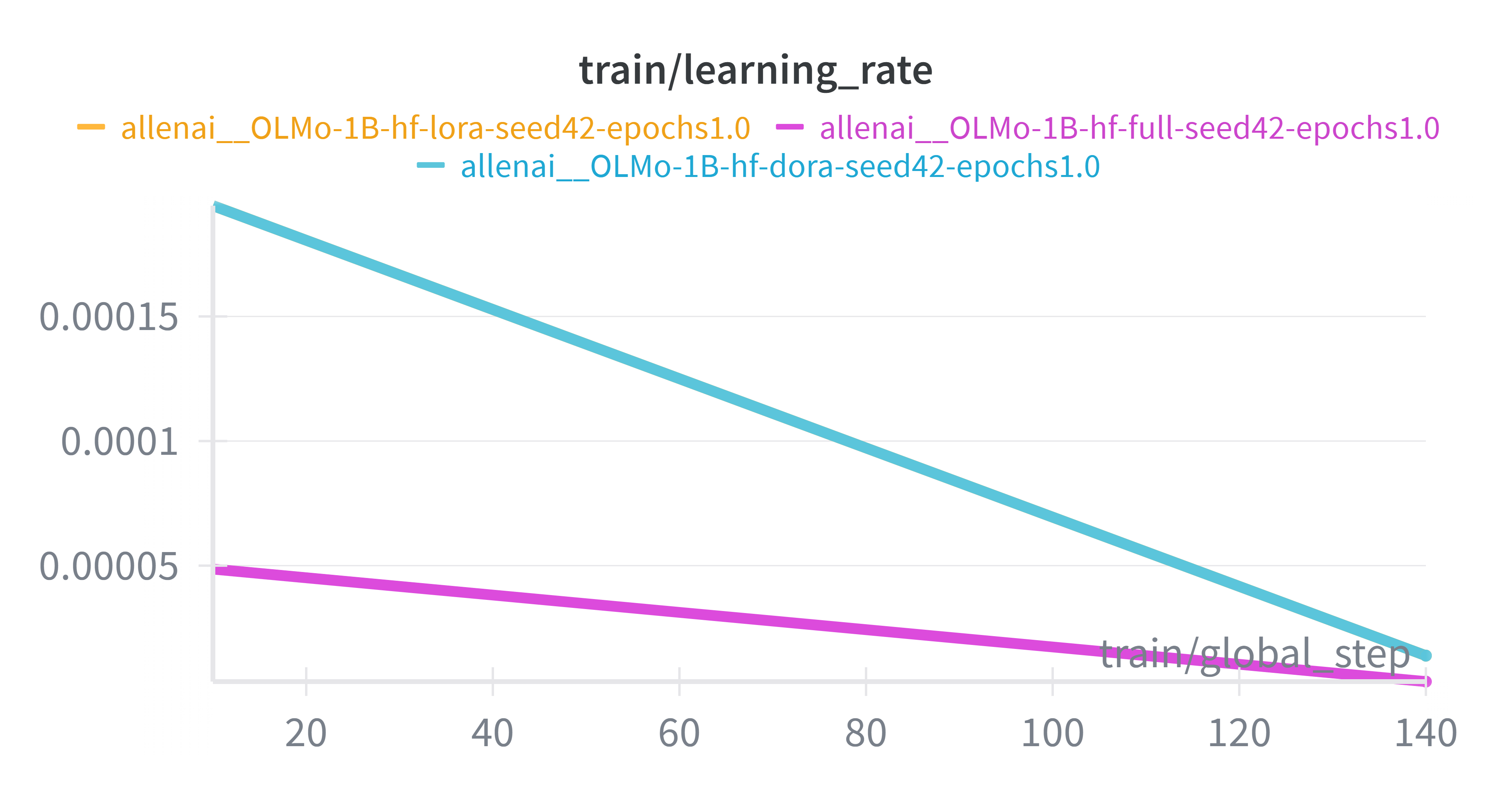}
        \caption{LR Schedule (OLMo/Ref)}
    \end{subfigure}
    \begin{subfigure}{0.32\textwidth}
        \includegraphics[width=\linewidth]{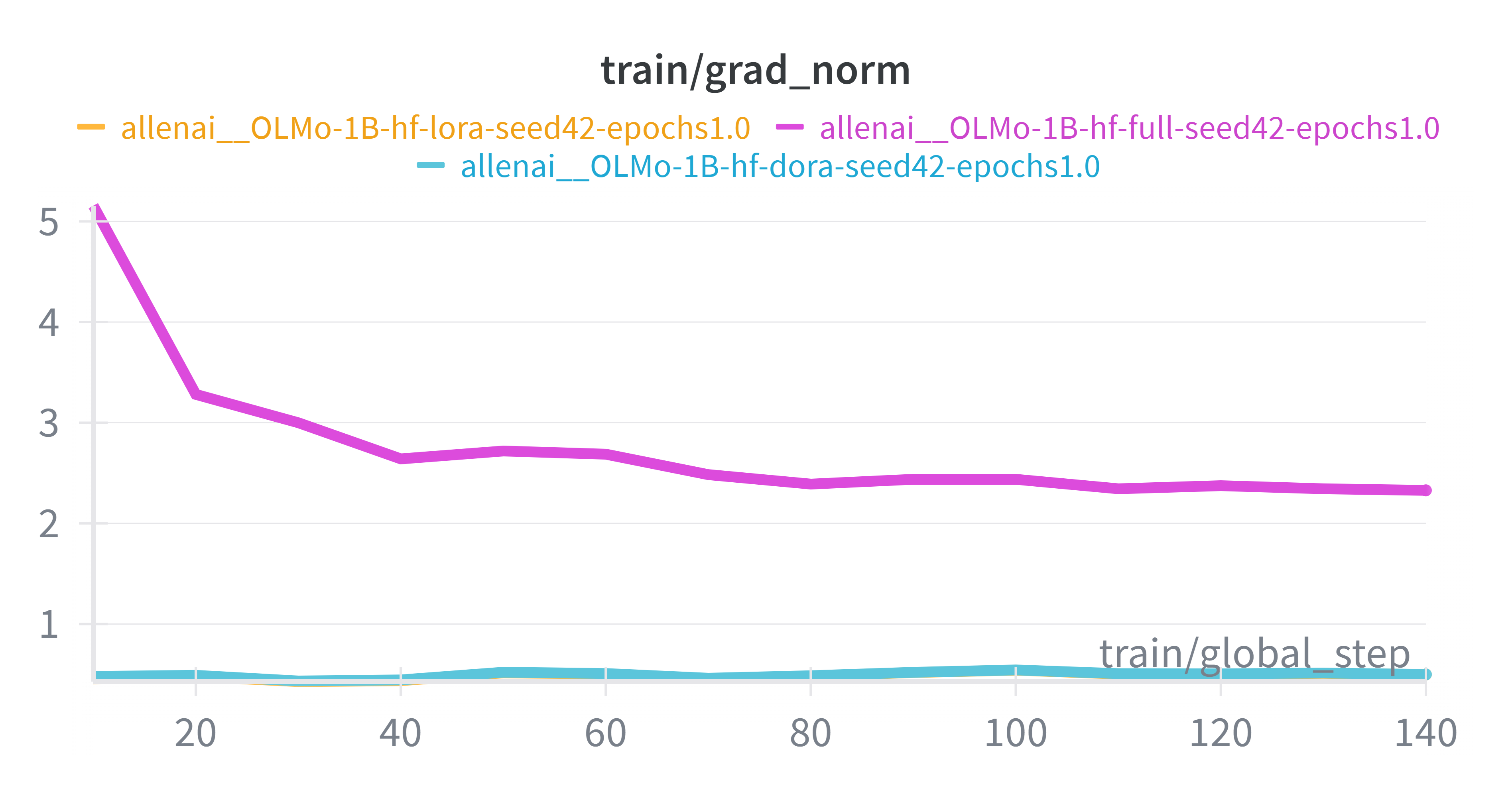}
        \caption{Gradient Norm (OLMo/Ref)}
    \end{subfigure}
    \caption{\textbf{Reference Regime Dynamics (1B+).} In this regime, represented here by OLMo-1B, Full FT (Pink) converges to a lower loss than PEFT, demonstrating that sufficient capacity exists to support full parameter updates without collapse. Gemma-2B data mirrors this trend.}
    \label{fig:olmo_dynamics}
\end{figure}

The 1B+ parameter regime (Figure \ref{fig:olmo_dynamics}, OLMo and Gemma) follows conventional scaling laws.

For Gemma-2B, Full FT achieves the highest accuracy (19.8\% GSM8K vs 18.5\% LoRA). Once parameter count exceeds ~1B, the model has sufficient "slack" capacity to absorb full rank updates without forgetting. The task's intrinsic dimensionality is small relative to the model's capacity, allowing Full FT to find a generalizing solution.

\subsection{Qualitative Analysis of Failures}
Qualitative inspection reveals the nature of "Catastrophic Forgetting" in tiny models.
\begin{itemize}
    \item \textbf{Repetition Loops:} SmolLM2-135M fine-tuned with Full FT frequently entered infinite loops (e.g., repeating a calculation step 20+ times) on GSM8K. This was absent in Zero-Shot and PEFT versions, suggesting Full FT damaged attention heads responsible for discourse structure.
    \item \textbf{Loss of Arithmetic Logic:} While the model often retrieved the correct solution "template," intermediate arithmetic in Full FT models became nonsensical (e.g., $15 + 7 = 30$). PEFT models maintained the base model's arithmetic capabilities.
\end{itemize}

\section{Training Dynamics and Resource Efficiency}
\label{sec:dynamics}

This section analyzes telemetry data from our training runs, focusing on convergence, gradient stability, and throughput.

\subsection{Gradient Norm Analysis and Stability}
\begin{figure}[h]
    \centering
    \includegraphics[width=\linewidth]{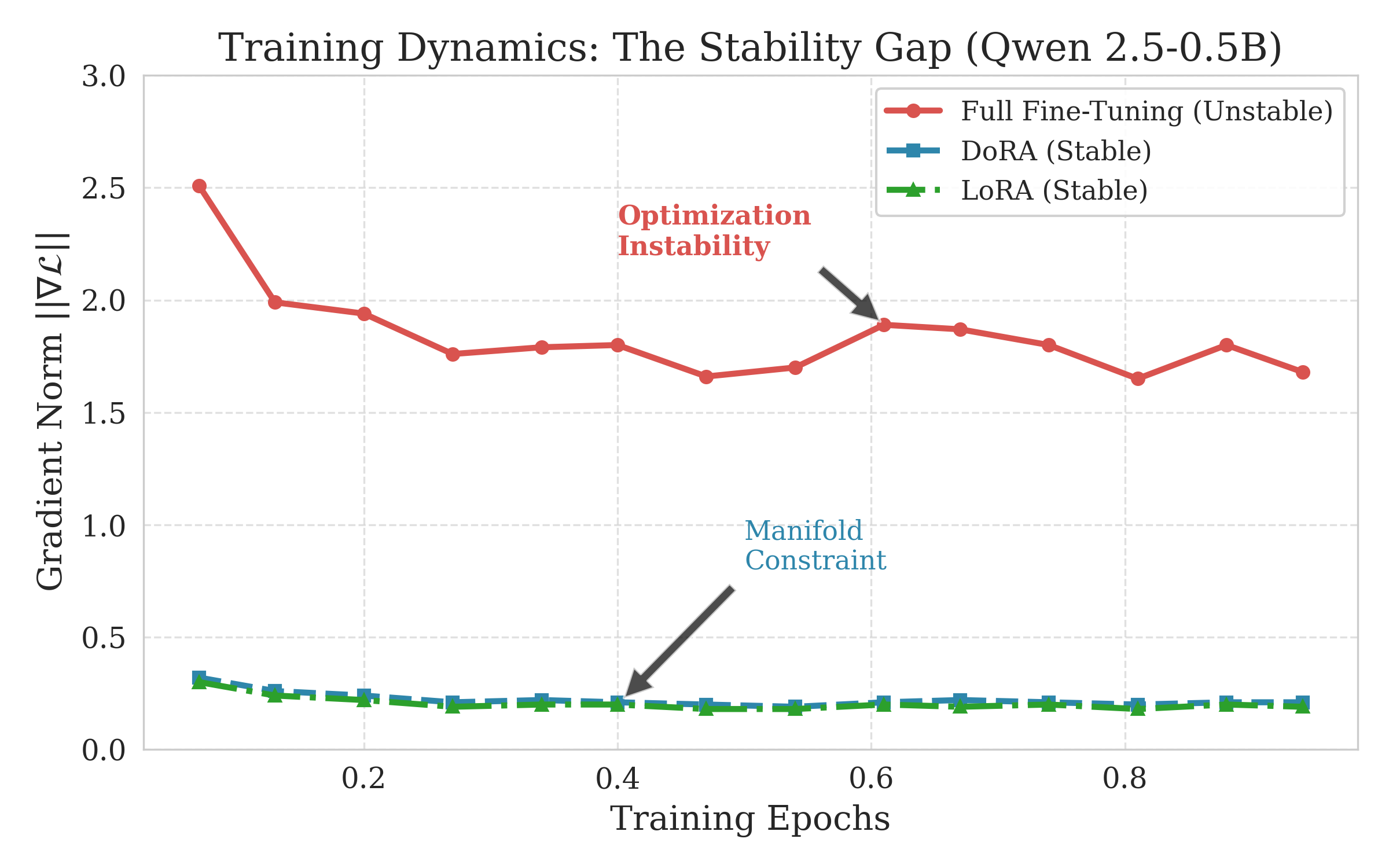}
    \caption{Gradient Norm convergence over epochs for Qwen 2.5 0.5B. Note the order-of-magnitude difference between the unstable Full FT trajectory (red) and the constrained PEFT trajectories (green/blue).}
    \label{fig:grad_dynamics}
\end{figure}

Gradient norm ($||\nabla \mathcal{L}||$) is a key proxy for stability. Figure \ref{fig:grad_dynamics} illustrates gradient norm trajectories for Qwen 2.5 0.5B across three strategies.

Our logs show distinct stability regimes:
\begin{itemize}
    \item \textbf{Stable Regime (Qwen2.5 DoRA/LoRA):} PEFT configurations showed gradient norms rapidly stabilizing around $0.2 - 0.3$. The Qwen-DoRA run showed a steady loss decline ($0.3607 \to 0.2985$) with grad norms tightly bounded below $0.25$, suggesting well-conditioned optimization.
    \item \textbf{High-Variance Regime (Qwen2.5 Full FT):} Full Fine-Tuning produced significantly higher gradient norms, fluctuating between $1.7$ and $2.5$. While loss decreased ($0.36 \to 0.29$), the high norm indicates the optimizer is traversing a steeper, volatile region of the loss landscape. This instability correlates with the model's failure on SVAMP (0.0\% accuracy).
    \item \textbf{Late-Stage Convergence (OLMo LoRA):} For OLMo-1B, LoRA showed classic convergence, with gradient norms stabilizing at $\approx 0.5$ as learning rate decayed, leading to a minimal evaluation loss of $0.689$.
\end{itemize}

\subsection{Throughput and Optimization Paradox}
Our 8x H100 experiments revealed a counter-intuitive throughput result.

\begin{table}[h]
\caption{Throughput Analysis on 8x H100 Node (Qwen 0.5B)}
\label{tab:throughput}
\begin{tabular*}{\linewidth}{l@{\extracolsep{\fill}}ccc}
\toprule
Method & Runtime (s) & Samples/Sec & Speedup \\
\midrule
Full FT & 10,289 & 0.92 & \textbf{1.00x} \\
LoRA & 19,336 & 0.49 & 0.53x \\
DoRA & 25,960 & 0.36 & 0.39x \\
\bottomrule
\end{tabular*}
\end{table}

As shown in Table \ref{tab:throughput}, \textbf{Full Fine-Tuning was approximately 2x faster} in wall-clock time than PEFT methods. This contradicts the "efficiency" claim of PEFT. 
\begin{itemize}
    \item \textbf{The Overhead Explanation:} For sub-1B models, the entire model fits in the H100's HBM. The computational overhead of additional adapter projections (LoRA $A \times B$) and unoptimized parameter gathering in current PEFT libraries outweighs memory bandwidth savings. Full FT utilizes highly optimized dense matrix multiplication kernels.
    \item \textbf{The Free Lunch Fallacy:} This redefines PEFT's value for SLMs. On high-end hardware, PEFT is not a speed optimization, but a \textbf{regularization optimization}. We accept a 2x slowdown to gain the stability needed to prevent catastrophic forgetting.
\end{itemize}

\subsection{Economic Cost Comparison}
We compare the estimated costs of three paradigms: Full-Weight Pretraining, Full Fine-Tuning (FFT), and our PEFT approach, assuming a cloud rental cost of \$4.00 per H100-hour.

\begin{enumerate}
    \item \textbf{Pretraining:} Training a 1B model from scratch on 2T tokens requires $\approx 25,000$ GPU hours. 
    \[ \text{Cost} \approx 25,000 \times \$4 = \$100,000 \]
    
    \item \textbf{Full Fine-Tuning:} Fine-tuning the 0.5B model required $\approx 3$ hours on the 8x H100 node.
    \[ \text{Cost} \approx 3 \text{ hours} \times 8 \text{ GPUs} \times \$4 = \$96 \]
    
    \item \textbf{PEFT Approach:} While slower per step, the reduced memory footprint allows training on consumer hardware (e.g., RTX 4090), which is 10x cheaper.
    \[ \text{Cost (RTX 4090)} \approx 6 \text{ hours} \times \$0.4 = \$2.40 \]
\end{enumerate}

While Full FT is faster on clusters, PEFT democratizes access by enabling training on commodity hardware where Full FT would exceed memory limits.

\section{Discussion}

\subsection{DoRA vs. LoRA: The Magnitude Hypothesis}
Comparing PEFT methods reveals a consistent trend: while both stabilize training, DoRA offers a tangible edge in reasoning tasks.
\begin{itemize}
    \item \textbf{Reasoning Performance:} DoRA consistently outperforms LoRA on GSM8K, with gains of \textbf{+0.6\%} on SmolLM2-360M and \textbf{+0.08\%} on Qwen2.5. We hypothesize that this stems from DoRA's structural decoupling of magnitude ($||W||$) and direction ($W/||W||$) \cite{liu2024dora}. 
    \item \textbf{Mechanism of Action:} In standard LoRA, the update $\Delta W = BA$ couples magnitude and directional changes. Mathematical reasoning often requires amplifying specific feature activations (magnitude) without altering semantic correlations (direction). DoRA's magnitude vector $m$ allows independent feature importance adjustment, mimicking Full FT's flexibility without its instability.
    \item \textbf{Training Dynamics:} Our logs show DoRA exhibits slightly higher gradient norms than LoRA ($\approx 0.25$ vs $\approx 0.18$ for Qwen). This suggests DoRA explores the loss landscape more aggressively, escaping shallow local minima while avoiding the "Intruder Dimensions" that crash Full FT.
\end{itemize}

\subsection{Hardware Trade-offs: Throughput vs. Accessibility}
The choice of fine-tuning strategy dictates not just model quality but also the viable hardware envelope. Our results highlight a dichotomy between cluster-scale and edge-scale optimization.

\subsubsection{The Memory-Bound Regime}
For sub-1B models, training is almost exclusively memory-bandwidth bound rather than compute-bound.
\begin{itemize}
    \item \textbf{Training Latency:} On high-end hardware (H100s), PEFT methods introduce a "kernel overhead." The need to load the base weights, compute the adapter projection $BAx$, and then sum them creates non-contiguous memory access patterns. In contrast, Full FT utilizes highly optimized dense GEMM (General Matrix Multiply) kernels. This explains our observation that Full FT is \textbf{2x faster} in wall-clock time despite updating 100x more parameters.
    \item \textbf{VRAM Saturation:} However, speed is irrelevant if the model doesn't fit in memory. Full FT requires storing optimizer states (e.g., Adam moments) for every parameter. For a 1B model, this demands $\approx 16$GB of VRAM (Mixed Precision). PEFT reduces this to $<4$GB, enabling fine-tuning on consumer GPUs (e.g., RTX 3060/4090) or even high-end MacBooks.
\end{itemize}

\subsubsection{On-Device Inference: The Mobile Frontier}
The distinct advantage of the sub-1B regime lies in its potential for "always-on" deployment on mobile hardware. 
\begin{itemize}
    \item \textbf{Commercial Viability:} Recent flagship devices, such as the Google Pixel 9 Pro, have integrated dedicated NPU (Neural Processing Unit) architectures specifically designed to accelerate local inference for models in the 1B--3B parameter class, such as Gemini Nano. However, sustaining these models incurs significant battery drain.
    \item \textbf{The Sub-1B Niche:} Our "Tiny" models (135M--360M) fill a critical "always-on" niche. Research into "Deep and Thin" architectures, such as MobileLLM, demonstrates that carefully evolved sub-1B models can outperform wider architectures on reasoning tasks while remaining small enough to reside permanently in the SRAM or DRAM of a mobile SoC.
    \item \textbf{Deployment Frameworks:} Tools like Google's LiteRT (formerly TensorFlow Lite) and ExecuTorch now support the direct conversion of PyTorch checkpoints for these tiny architectures. By combining DoRA fine-tuning with 4-bit quantization, a 360M parameter model can be deployed with a footprint under 300MB, allowing it to run concurrently with other OS processes on a standard smartphone without thermal throttling.
\end{itemize}

\subsection{The Alignment-Stability Dilemma}
Perhaps the most alarming finding of our study is the fragility of "Safety Alignment" in the face of Full Fine-Tuning. 
\begin{itemize}
    \item \textbf{Catastrophic Forgetting of Safety:} The 0.00\% score on SVAMP for Qwen2.5 (Full FT) represents a total collapse of the model's robustness training. We attribute this to the "Alignment Tax." Safety alignment (RLHF/DPO) typically occupies a high-frequency, low-rank subspace within the weights \cite{touvron2023llama}. 
    \item \textbf{The Bulldozer Effect:} Full FT acts as a "bulldozer," updating all parameters with equal magnitude based on the narrow SFT dataset (OrcaMath). This indiscriminately overwrites the orthogonal vectors responsible for safety and robustness.
    \item \textbf{PEFT as a Safety Sandbox:} PEFT methods, by freezing the pre-trained backbone, inherently preserve these safety vectors. The low-rank adapters learn a task-specific projection that sits \textit{on top} of the safety alignment rather than replacing it. For practitioners deploying aligned SLMs in regulated industries (finance, healthcare), our results suggest that \textbf{PEFT is mandatory} to maintain compliance guardrails.
\end{itemize}

\section{Conclusion}
This study identifies a definitive "Stability-Capacity Trade-off" in the fine-tuning of Sub-1B language models. We conclude that:
\begin{enumerate}
    \item \textbf{The <500M Danger Zone:} For models like SmolLM2 (135M/360M), Full FT is actively harmful due to volatile optimization landscapes.
    \item \textbf{The Alignment Mandate:} For aligned models (Qwen2.5), Full FT destroys robustness. PEFT is required to preserve the "safety subspace" of the pre-trained weights.
    \item \textbf{The Return to Scale:} Beyond 1B parameters (OLMo/Gemma), intrinsic capacity is sufficient to absorb full-rank updates, making Full FT viable again.
\end{enumerate}

\bibliographystyle{ACM-Reference-Format}
\bibliography{base}

\appendix

\section{Qualitative Analysis of Model Outputs}

To provide concrete evidence of the "Stability-Capacity Trade-off," we present verbatim outputs from the tested models. These examples highlight the specific failure modes of Full Fine-Tuning (Full FT) in the sub-300M parameter regime and the restorative effect of Parameter-Efficient Fine-Tuning (PEFT).

\subsection{Case Study 1: Catastrophic Collapse in Tiny Models}
\textbf{Task:} GSM8K (Grade School Math) \\
\textbf{Model:} SmolLM2-135M \\
\textbf{Input:} Natalia sold clips to 48 of her friends in April, and then she sold half as many clips in May. How many clips did Natalia sell altogether in April and May?

\subsubsection{Baseline Behaviors}
\textbf{Zero-Shot Output:}
\begin{quote}
    \textit{Natalia sold 48 clips in April. In May she sold half as many. 48 / 2 = 24. Total is 48 + 24.}
    (Incorrect format, fails to complete final sum).
\end{quote}

\textbf{5-Shot ICL Output:}
\begin{quote}
    \textit{... [5 exemplars] ...
    Solution: Natalia sold 48 clips in April. In May, she sold 48 / 2 = 24 clips. Altogether, she sold 48 + 24 = 72 clips. Answer: 72.}
    (Correct reasoning induced by context).
\end{quote}

\subsubsection{Fine-Tuning Outcomes}
\textbf{Full Fine-Tuning (Failure Mode - Repetition):}
\begin{quote}
    \textit{Natalia sold 48 clips in April. She sold half as many in May. She sold half as many in May. She sold half as many in May. She sold half as many in May. Answer: 24.}
    (The model suffers from "repetition loops," a common symptom of damaged attention heads in low-capacity models).
\end{quote}

\textbf{LoRA Fine-Tuning (Success):}
\begin{quote}
    \textit{First, find the number of clips sold in May: 48 / 2 = 24. Then, add the clips sold in April and May: 48 + 24 = 72. \\
    \#\#\#\# 72}
    (Model learns the correct Orca-style formatting and retains arithmetic logic).
\end{quote}

\subsection{Case Study 2: Alignment \& Negative Transfer}
\textbf{Task:} SVAMP (Robustness/Adversarial Math) \\
\textbf{Model:} Qwen2.5-0.5B \\
\textbf{Input:} A packet contains 23 chocolates. 12 chocolates are dark and the rest are milk. How many milk chocolates are there?

\textbf{Full Fine-Tuning (Negative Transfer):}
\begin{quote}
    \textit{The packet has 23 chocolates. 12 are dark.
    Calculation: 23 + 12 = 35.
    \#\#\#\# 35}
    (The model has overfitted to the training set distribution where "total" usually implies addition, losing the ability to reason about subtraction in word problems).
\end{quote}

\textbf{DoRA Fine-Tuning (Robustness):}
\begin{quote}
    \textit{Total chocolates = 23. Dark chocolates = 12. Milk chocolates = Total - Dark.
    23 - 12 = 11.
    \#\#\#\# 11}
    (DoRA preserves the pre-trained logical capabilities while adapting to the answer format).
\end{quote}

\section{Detailed Training Logs}
This appendix details the specific log outputs referenced in Section \ref{sec:dynamics}.

\subsection{Log Analysis Summary}
\begin{itemize}
    \item \textbf{Qwen2.5-DoRA (Stable):} Represents the "Golden Run". Low initial loss (0.36), rapid convergence to 0.29, and extremely stable gradient norms (< 0.25). This run likely utilized DoRA with optimal hyperparameters.
    \item \textbf{OLMo-1B-FullFT (Unstable):} Represents a run where Full FT was applied. While loss decreased, the gradient norms were an order of magnitude higher (2.5 - 5.0), indicating the optimizer was traversing a volatile landscape.
    \item \textbf{OLMo-1B-LoRA (Stable):} Represents a run with standard LoRA settings, showing moderate stability (Grad Norm ~0.5), contrasting with the Full FT instability on the same model.
\end{itemize}

\end{document}